\documentclass[letterpaper]{article} 
\usepackage{aaai2026}  
\usepackage{times}  
\usepackage{helvet}  
\usepackage{courier}  
\usepackage[hyphens]{url}  
\usepackage{graphicx} 
\usepackage{natbib}  
\usepackage{caption} 
\usepackage{algorithm}
\usepackage{algorithmic}
\usepackage{multirow}
\usepackage{array}
\usepackage{booktabs}  

\usepackage{newfloat}
\usepackage{listings}
\DeclareCaptionStyle{ruled}{labelfont=normalfont,labelsep=colon,strut=off} 
\floatstyle{ruled}
\newfloat{listing}{tb}{lst}{}
\floatname{listing}{Listing}
\title{Hybrid Hypergraph Networks for Multimodal Sequence Data Classification }
\author {
    Feng Xu\textsuperscript{\rm 1,2}, 
    Hui Wang\textsuperscript{\rm 1},
    Yuting Huang\textsuperscript{\rm 3},
    Danwei Zhang\textsuperscript{\rm 4},
    Zizhu Fan\textsuperscript{\rm 5*}
}
\affiliations {
    \textsuperscript{\rm 1}East China Jiaotong University, Nanchang, China\\
    \textsuperscript{\rm 2}Quzhou College of Technology, Quzhou, China\\
    \textsuperscript{\rm 3}Zhejiang University, Hangzhou, China\\
    \textsuperscript{\rm 4}The State Key Laboratory of Synthetical Automation for Process Industries,
Northeastern University, Shenyang, China\\
    \textsuperscript{\rm 5}College of Computer Science and Technology, Shanghai University of Electric Power, Shanghai, China
\\
    \small \texttt{fengxu15@mails.jlu.edu.cn, huiwangens@163.com, yutinghuang@zju.edu.cn} \\
    \small \texttt{1910253@stu.neu.edu.cn, zzfan3@shiep.edu.cn}
}

\usepackage{bibentry}

\usepackage{enumitem}

\begin{document}

\maketitle

\renewcommand{\thefootnote}{\fnsymbol{footnote}}
\footnotetext[1]{Corresponding author.}

\begin{abstract}

Modeling temporal multimodal data poses significant challenges in classification tasks, particularly in capturing long-range temporal dependencies and intricate cross-modal interactions. Audiovisual data, as a representative example, is inherently characterized by strict temporal order and diverse modalities. Effectively leveraging the temporal structure is essential for understanding both intra-modal dynamics and inter-modal correlations. However, most existing approaches treat each modality independently and rely on shallow fusion strategies, which overlook temporal dependencies and hinder the model's ability to represent complex structural relationships. To address the limitation, we propose the hybrid hypergraph network (HHN), a novel framework that models temporal multimodal data via a segmentation-first, graph-later strategy. HHN splits sequences into timestamped segments as nodes in a heterogeneous graph. Intra-modal structures are captured via hyperedges guided by a maximum entropy difference criterion, enhancing node heterogeneity and structural discrimination, followed by hypergraph convolution to extract high-order dependencies. Inter-modal links are established through temporal alignment and graph attention for semantic fusion. HHN achieves state-of-the-art (SOTA) results on four multimodal datasets, demonstrating its effectiveness in complex classification tasks.

\end{abstract}


\section{Introduction}

In real-world scenarios, multimodal data often exhibit temporal characteristics, with significant heterogeneity across modalities and strong temporal dynamics, posing higher demands on event modeling and classification accuracy \cite{zhou2025dynamic,zhao2024deep,yilmaz2021multimodal}. Meanwhile, temporal data typically possess inherent graph-structured properties \cite{rossi2020temporal,yu2017spatio}, which necessitate modeling both temporal evolution and complex inter-modal relationships. Relying solely on a single modality is insufficient to effectively capture the cross-temporal and cross-modal dependencies present in multimodal temporal data.

\begin{figure}[t]
\centering
\includegraphics[width=0.9\columnwidth]{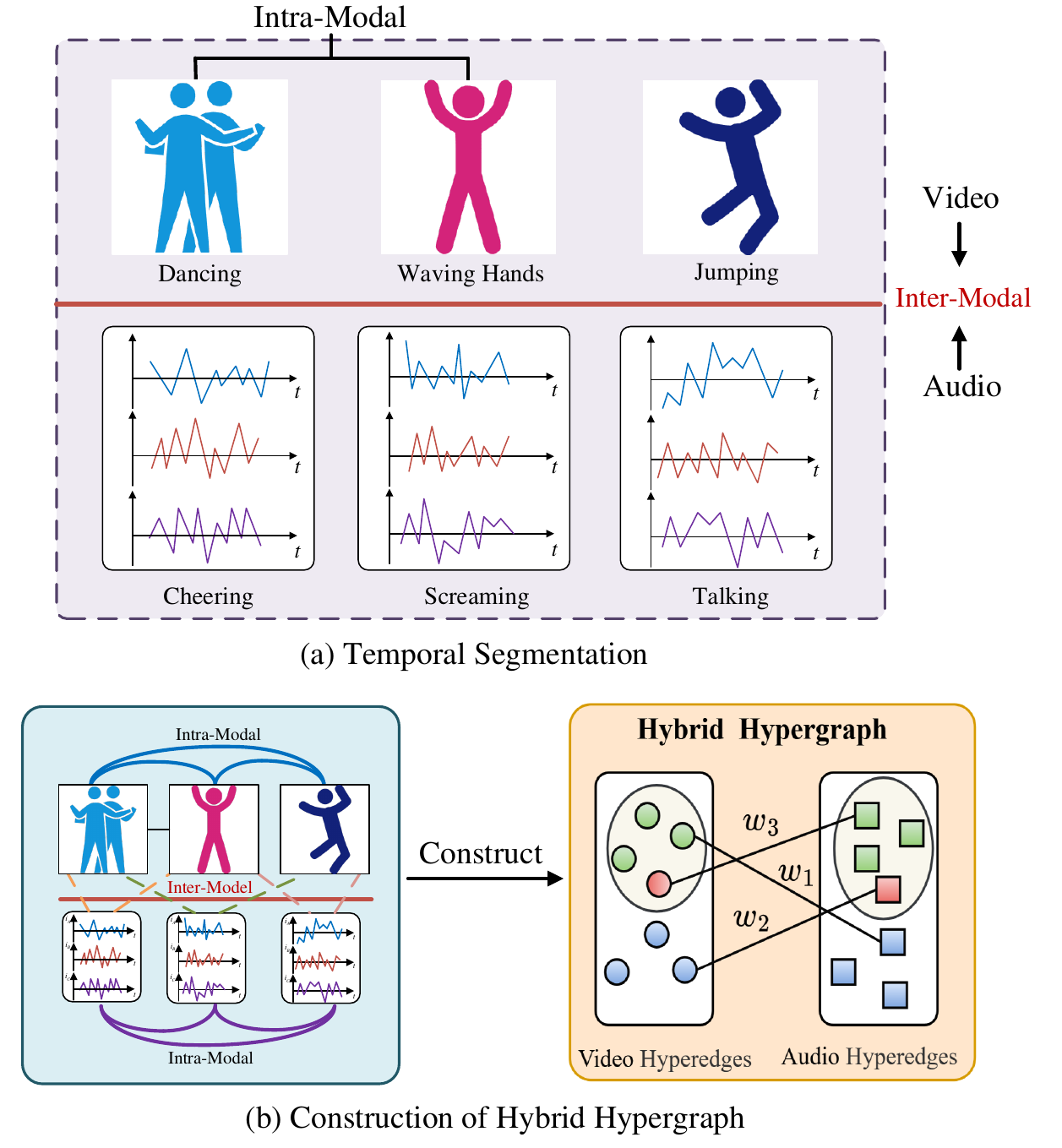}
\caption{Temporal segmentation and construction of the temporal hybrid hypergraph for audiovisual signals in a concert}

\label{fig1_dancing}
\end{figure}

As illustrated in Figure~\ref{fig1_dancing}, the audiovisual scenario demonstrates the significant complementary role of the visual modality in semantic enhancement, which helps improve the model's ability to discriminate complex temporal events. Multimodal data, exemplified by audiovisual sequences, inherently exhibit strict temporal order and modal diversity. Therefore, effectively modeling their temporal structures is crucial for capturing both intra-modal dynamics and inter-modal interactions.

Existing multimodal approaches typically adopt modality-specific modeling followed by shallow feature fusion, where information is integrated through concatenation, attention mechanisms, or loss constraints \cite{yang2022interpolation,saeed2021contrastive}. However, such methods struggle to capture temporal alignment across modalities and long-range structural dependencies effectively. In recent years, researchers have increasingly explored representing temporal multimodal relationships using unified graph structures, where continuous segments are treated as nodes and connections are constructed to model both intra-modal and inter-modal structural dependencies, thereby enabling the capture of dynamic relationships that evolve \cite{ran2025fine,liu2025audio}.

Although multimodal graph modeling has made progress, several challenges remain. Many approaches rely on fixed rules such as k-nearest neighbors for edge construction, which limits their ability to capture structural heterogeneity and semantic differences. Inadequate temporal alignment across modalities further hampers dynamic relation modeling. These issues are especially pronounced in tasks such as audiovisual event recognition and industrial monitoring, where data exhibit strong temporal variation and modal heterogeneity. Therefore, an adaptive graph framework is needed to enable flexible edge construction, heterogeneous dependency modeling, and temporal alignment for improved structural representation and dynamic fusion of multimodal sequences.

To address the challenges, we propose the hybrid hypergraph network (HHN) for modeling multimodal sequential data. HHN integrates the high-order representation capability of hypergraphs with the adaptive edge construction of graph attention mechanisms, enabling more effective modeling of complex structural dependencies. The overall strategy follows a ``partition-then-construct'' paradigm, where sequential data are divided into timestamped segments that serve as nodes in a heterogeneous graph.

For intra-modal modeling, a hyperedge connection mechanism based on the principle of maximum difference entropy is proposed to capture structural heterogeneity among nodes. The temporal window size is adaptively adjusted according to the difference between each node's entropy and the global average entropy, enabling high-entropy nodes to establish connections with structurally diverse neighborhoods. The integration of this entropy-guided connection strategy with hypergraph convolution operations enables effective modeling of high-order dependencies and local temporal variations within each modality. For inter-modal modeling, graph attention networks (GATs) are introduced to perform temporal alignment and cross-modal edge construction. The attention-based architecture facilitates identification of critical cross-modal connections and enhances the model's capacity to capture long-range dependencies. In addition, the hypergraph paradigm demonstrates strong potential in the analysis of temporal multimodal data beyond traditional graph tasks.

The main contributions are as follows:

\begin{itemize}

\item We propose HHN, a framework that leverages hypergraphs to model intra-modal temporal dependencies in multimodal sequential data, overcoming the limitations of pairwise graph structures in capturing high-order relationships.

\item We design an entropy-guided adaptive window mechanism for hyperedge construction, enabling robust modeling of structural heterogeneity and long-range temporal dependencies.

\item Our experiments on AudioSet, AVE, RAVDESS, and FMF demonstrate that HHN achieves state-of-the-art (SOTA) performance in multimodal temporal sequence classification tasks.
\end{itemize}

\section{Related Works}
\label{sec: relation works}
We briefly review related works on multimodal graph learning and temporal graph learning, which together support the modeling of both cross-modal relationships and temporal dynamics in our proposed multimodal graph-based framework.

\subsection{Multimodal Graph Learning}

Graph-based multimodal learning has received growing attention for its ability to model structured relationships among heterogeneous data sources. Researchers have proposed various graph construction strategies. Zhu et al \cite{zhu2025mosaic} present Mosaic, a benchmark that standardizes graph construction procedures and modality integration approaches, providing a unified evaluation framework for multimodal learning. With the increasing application of graph structures, graph neural networks (GNNs)~\cite{corso2024graph,wu2020comprehensive} have been widely adopted to model more complex temporal and semantic relationships.

Beyond structural modeling, recent efforts have also focused on enhancing semantic alignment and representation learning. Li et al.~\cite{li2025kg4mm} conduct a detailed survey on knowledge-enhanced multimodal learning and highlight the role of knowledge graphs in providing explicit semantic grounding and improving model interpretability. Xia et al.~\cite{xia2023graph} further explore contrastive learning in multimodal scenarios by aligning modality representations through graph embeddings, thereby improving clustering performance. Although these studies advance the field regarding structure, semantics, and representation alignment, challenges remain in modeling fine-grained temporal dependencies and high-order cross-modal interactions, particularly in sequential or video-based data.

\subsection{Temporal Graph Learning}

Temporal graph neural networks (TGNNs) incorporate temporal information as an important weight in constructing the graph neural network model, capturing different levels of temporal evolution characteristics. Currently, temporal graph learning methods encode time information as discrete time steps or continuous timestamps, enabling the model to capture temporal dependencies with different levels of granularity. By integrating temporal information, temporal graph learning can effectively analyze the dynamic evolution of graph data and shows broad potential in various fields, such as traffic flow prediction~\cite{xu2023dynamic}, epidemic simulation~\cite{gao2021stan}, and energy consumption forecasting~\cite{shi2024stgnets}.

Although temporal graph methods have achieved notable success in downstream graph-related tasks, their broader application in complex multimodal scenarios remains underexplored \cite{liu2024self,liu2022embedding}. In such settings, temporal structures are essential for capturing both intra-modal dynamics and inter-modal interactions. However, it remains challenging to effectively model high-order relationships within each modality, particularly the interactions among multiple factors that jointly influence node states under complex temporal conditions.

To tackle this issue, we propose a hybrid construction strategy based on hypergraphs, where hyperedges connect multiple nodes to explicitly encode high-order structural dependencies. This approach enhances the expressive capacity of temporal graphs by capturing relational patterns beyond simple pairwise links. Moreover, we integrate hypergraph convolutional networks (HGNNs) \cite{gao2022hgnn+,feng2019hypergraph} into the temporal modeling framework, enabling more flexible and effective representation of complex node relations and modality interactions. Compared to conventional graph neural networks, the proposed method provides key advantages, including the ability to model non-pairwise interactions, reduce information loss during neighborhood aggregation, and more accurately capture high-order structural patterns in multimodal temporal data.

\section{Methods}

\begin{figure*}[!t]
\centering
\hspace*{2pt}
\includegraphics[width=0.9\textwidth]{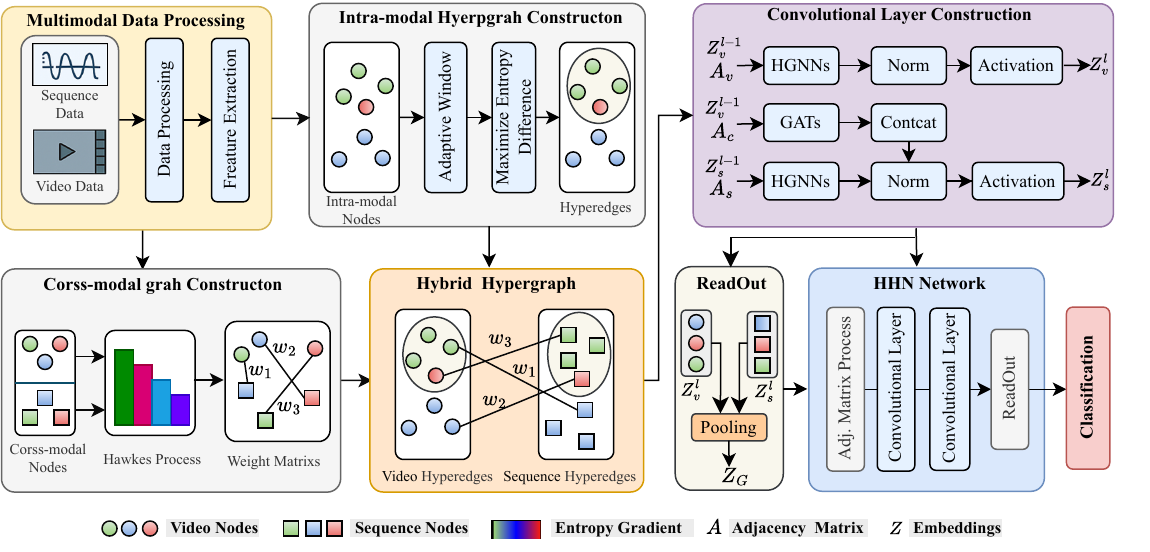} 
\caption{The overall framework of HHN.
We construct a hybrid hypergraph from multimodal sequence and video data with two node types and hyperedges. Intra-modal hypergraphs and cross-modal graphs are built using adaptive windowing, Hawkes processes, and entropy-based weighting. Sequence and video nodes are processed by separate HGNNs, followed by cross-modal aggregation via GATs. Final node embeddings are pooled by a ReadOut module for classification.}
\label{fig2}
\end{figure*}

The methodology of HHN is structured around four core components: feature extraction, intra-modal graph construction, inter-modal graph construction, and hybrid convolution. The overall framework is depicted in Figure~\ref{fig2}, and the associated mathematical notations are detailed in Table~\ref{table:notation}.

\subsection{Definition of the Hybrid Hypergraph}

To express the relationship between intramodal and cross-modal in graph convolution, we propose a new method for constructing a time-multimodal hybrid graph. This architecture consists of two parts: intra-modal, represented by a hypergraph to capture complex relationships within a single modality, and cross-modal, represented by a weighted ordinary graph, where temporal relationships are reflected through weights. Time information is integrated into the multimodal hybrid graph construction process by capturing dynamic relationships that change over time within each modality. The definition of the hybrid graph is as follows:

\begin{equation}
{G_m} = ({V_m},{E_t},{E_{h}},{\cal T},{\cal F}) \label{eqn:eq4}
\end{equation}
Where $V_m$ is the set of nodes shared by the temporal graph and the hypergraph. $E_t$ is the set of edges in the temporal graph, representing the direct sequential relationships in the time series. $E_h$ is the set of hyperedges in the hypergraph, comprising subsets of nodes used to describe complex multivariate relationships. $\cal T$ is the set of timestamps, which can denote the edges' temporal attributes. $\cal F$ is the set of node features, with each node's features shared by both the temporal graph and the hypergraph.

\begin{table}[h]
\centering
\renewcommand{\arraystretch}{1.2}  
\begin{tabular}{c m{5.8cm}}  
\hline
\multicolumn{1}{c}{Notation} & \multicolumn{1}{c}{Description} \\
\hline
$G_m$ & Hybrid graph: $(V_m, E_t, E_h, \mathcal{T}, \mathcal{F})$ \\
$V_m$ & Set of multimodal nodes \\
$E_t$, $E_h$ & Edge set of temporal graph and hypergraph \\
$\mathcal{T}$, $\mathcal{F}$ & Timestamps and feature vectors of nodes \\
$H(v_i)$ & Entropy of node $v_i$ \\
$Z_m^l$ & Embedding of modality $m$ at layer $l$ \\
$P_v$, $P_s$ & Learnable aggregation weights for video and sequence modalities \\
$A_s$, $A_v$, $A_c$ & Adjacency matrices for sequence, video, and cross-modal graphs \\
\hline
\end{tabular}
\caption{Notation Description}
\label{table:notation}
\end{table}

\subsection{Feature Extraction}

\textbf{Sequence Data Encoding:} To extract features from audio or current data, the raw data is divided into fixed-length segments with overlapping regions between consecutive segments. For audio data, each segment has a duration of 960 milliseconds and an overlap of 764 milliseconds. A log-mel spectrogram is computed for each segment using the short-time Fourier transform, and the resulting spectrogram is fed into the pre-trained VGGish \cite{hershey2017cnn} to extract a 128-dimensional feature vector. The segmentation process for current data follows the same scheme as audio data, but feature extraction is performed using TS2Vec \cite{yue2022ts2vec}, which maps each segment to a 128-dimensional feature vector. 

\textbf{Video Encoding:} To extract features from video nodes, we divide the video into non-overlapping segments of 250 milliseconds. Each segment is then processed using the S3D \cite{xie2018rethinking}, which has been trained through self-supervised learning, generating a 1024-dimensional feature vector.

\subsection{Intra-Modal hypergraph Construction}~\label{sec:infra_modal_hypergraph}

Entropy, a fundamental concept in information theory, measures the uncertainty of a system and is widely used to characterize the diversity and distribution of information within data \cite{shannon1948mathematical}. In graph-based representations, a node’s entropy quantifies how uniformly its feature values are distributed across dimensions, serving as an indicator of the node’s information richness and uncertainty~\cite{fixelle2025hypergraph,luo2021graph}. Leveraging this property in hybrid hypergraph construction, maximizing entropy differences between nodes promotes feature heterogeneity, which in turn enhances the model’s ability to capture complex multimodal dependencies.

\begin{figure}[t]
\centering

\includegraphics[width=0.9\columnwidth]{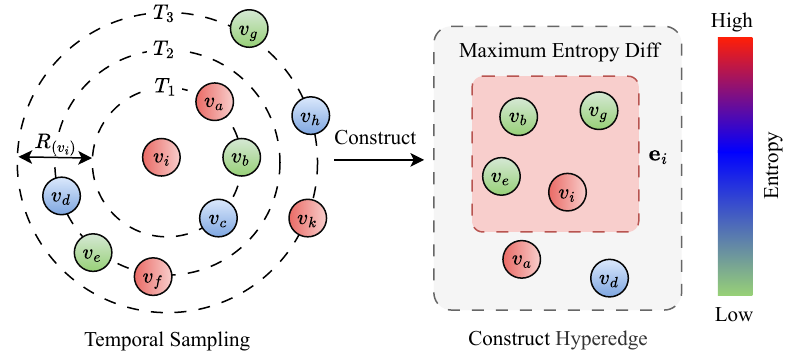}
\caption{
The diagram illustrates entropy-guided hyperedge construction, where three nodes with the highest entropy differences from $v_i$ are selected within the adaptive window $R_{(v_i)}$ to form a hyperedge.
}
\label{fig1}
\end{figure}

While entropy provides a theoretical foundation for measuring node uncertainty, constructing effective hyperedges based on it remains challenging. In scenarios with a small number of nodes, fully connected hypergraphs may seem reasonable but often introduce redundant links and weaken structural discriminability. We propose a construction strategy based on maximizing entropy difference, which selects node combinations that form sparse yet diverse hyperedges. The approach aligns with the Shannon sampling theorem and information-theoretic representation learning methods \cite{sun2020infograph}, ensuring that each hyperedge captures complementary and discriminative information beneficial for downstream classification tasks. The formal definition of the entropy of a node is as follows \cite{shannon1948mathematical}:

\begin{equation} H(v_i) = - \sum_{j=1}^{m} P(f_j) \log P(f_j) \label{eqn:entropy} \end{equation}
where $H(v_i)$ represents the entropy of node $v_i$, $m$ is the number of features for the node, $f_j$ denotes the $j$-th feature of node $v_i$, and $P(f_j)$ represents the probability of feature $f_j$ for node $v_i$. The entropy quantifies the uncertainty or variability in the feature distribution of the node.

In the intra-modal graph construction, we propose an adaptive window-based hyperedge generation strategy, where the window size is dynamically determined by each node's feature entropy. This entropy-guided mechanism allows nodes with higher uncertainty to form larger neighborhoods, enabling them to capture more complex local patterns. Here, the term adaptive window function refers to this entropy-based neighborhood scaling strategy. Formally, the entropy-adaptive range \( R(v_i) \) for node \( v_i \) is computed as:

\begin{equation} R(v_i) = \left\lfloor R_{\min} + \alpha \cdot \frac{H(v_i)}{\bar{H}} \right\rfloor \label{eqn:window} \end{equation}
where $H(v_i)$ is the entropy of node $v_i$, $\bar{H}$ is the global average entropy, $R_{\min}$ is the minimum window size, and $\alpha$ is a coefficient controlling the adjustment of the window size.

As illustrated in Figure~\ref{fig1}, given the adaptive window \( R(v_i) \) of node \( v_i \), we aim to select \( n \) nodes to form a hyperedge \( \mathbf{e}_i \) with \( v_i \), maximizing the sum of entropy differences between the selected nodes and \( v_i \), subject to \( |\mathbf{e}_i| = n \) and \( \mathbf{e}_i \in E_h \). 
The optimization objective is formulated as: 

\begin{equation}
 \arg\max_{\mathbf{e}_i\subset R(v_i)} \sum_{v_j \in \mathbf{e}_i} \left| H(v_i) - H(v_j) \right|
\label{eqn:entropy_diff}
\end{equation}

\subsection{Cross-modal Graph Construction}
 GATs are used to construct a weighted ordinary graph in the cross-modal part. The weights in the adjacency matrix are adjusted to reflect the temporal correlations between nodes. To incorporate the temporally correlated edge set $E_t$ into the definition of the hybrid hypergraph, we can define $E_t$ as:

\begin{equation}
E_t = \{(v_i, v_j) \mid A_{i,j} W_{i,j} > 0\}
\label{eqn:E_t}
\end{equation}
where $A_{i,j}$ represents the combined weight in the final adjacency matrix between nodes $v_i$ and $v_j$, $W_{i,j}$ is the temporal weight reflecting the temporal correlation between $v_i$ and $v_j$.

To reflect the impact of temporal factors on the relationships between nodes more accurately, we assign higher weights to neighbors with smaller time intervals. Thus, for each neighbor of a node, temporal weights determined by the Hawkes process \cite{hawkes1971point} are used to weigh the edges. The weighting formula can be expressed as:

\begin{equation}
{W_{i,j}} = \exp( - {\rm{ }}\frac{{{t_{max}} - {t_i} + 1}}{{{t_{max}} - {t_{min}} + 1}}),{\rm{  where \ j}} \in {N_i} \label{eqn:eq7}
\end{equation}
where $t_i$ represents the timestamp of node $i$; $j$ represents the neighbor nodes of node $i$; $t_{max}$ and $t_{min}$ represent the maximum and minimum timestamps in the set, respectively.

\subsection{Convolutional Layer Construction}

To enable effective inter-modal fusion, HHN constructs cross-modal edges based on temporal alignment and semantic similarity, followed by a graph attention mechanism to integrate complementary information across modalities. This design allows the model to capture fine-grained temporal correlations and cross-modal dependencies in a unified framework. For intra-modal representation, HHN constructs hyperedges via entropy-based segmentation, capturing high-order structures within each modality. Hypergraph convolution is then applied to enhance node representations:

\begin{equation}
Z_m^l = {{HGNNs}}({A^m},Z_m^{l - 1}) \label{eqn:eq8}
\end{equation}
where $A^m$ is the hypergraph adjacency matrix of the modality; $Z_m^{l - 1}$ is the embedding from the previous layer.

We employ hypergraph convolution and GATs to model interactions between video and sequence modalities. GATs enable adaptive feature alignment across modalities, enhancing semantic consistency and reinforcing inter-modal dependencies. Video features are processed via HGNNs and dynamically fused with sequence features through attention-based aggregation, forming the joint representation for the next layer. The process is formally expressed as:

\begin{equation}
Z_s^l = HGNNs(A_s,\ Z_s^{l - 1}\ \Vert \ GATs(A_c,\ Z_v^{l - 1})) \label{eqn:eq9}
\end{equation}

\begin{equation}
{Z_{\rm{v}}}^l = HGNNs({A_v},{Z_v}^{l - 1})\label{eqn:eq10}
\end{equation}
where $Z_v$ and $Z_s$ are the embedding vectors for the video and sequence modalities, respectively; ${Z_v}^{l - 1}$ and ${Z_s}^{l - 1}$ are their corresponding representations from the previous layer; $A_v$ and $A_s$ denote the hypergraph adjacency matrices capturing intra-modal relationships within the video and sequence modalities, respectively; $A_c$ represents the adjacency matrix modeling inter-modal interactions between the two modalities; and $\Vert$ denotes the concatenation of the two inputs along the feature dimension.

To form a comprehensive representation, the features from all modalities are integrated using a unified merging strategy. This strategy is implemented by the following aggregation function, which considers the temporal alignment and interdependencies between the modalities:
\begin{equation}
{Z_{G}} = {\rm{ReadOut(G_i)}}= {Z_v}^l{P_v} + {Z_s}^l{P_s} \label{eqn:eq11}
\end{equation}
where $Z_G$ is the final graph embedding, ${Z_v}^l$ and ${Z_s}^l$ are the embeddings of the video and sequence modalities at layer $l$, $P_v$ and $P_s$ are learnable aggregation vectors used to balance the contributions of different modalities at layer $l$, and $h$ is the aggregation function.

The procedure of the temporal multimodal hybrid graph fusion network is shown in Algorithm \ref{alg:TMHGFN}, and the network is trained using the focal loss $L$ as follows:

\begin{equation}
L =  - \sum\limits_n {{{(1 - {{\hat y}_n})}^{2} }} \log ({\tilde y_n}) \label{eqn:eq12}
\end{equation}
where ${\hat y}_n$ denotes predicted labels, $\tilde y_n$ denotes ground-truth labels.


\begin{algorithm}[tb]
\caption{HHN Algorithm}
\label{alg:TMHGFN}
\textbf{Input}: Video clips, sequence data segments\\
\textbf{Output}: Predicted labels for the video clips and sequence data segments
\begin{algorithmic}[1] 
\STATE Construct ${G_m} = ({V_m},{E_t},{E_{h}},{\cal T},{\cal F})$
\STATE Batch the data and initialize model parameters
\FOR{each epoch}
    \FOR{each batch}
        \STATE Construct the hypergraph based on Eq.~(\ref{eqn:entropy_diff})
        \STATE Calculate the weights of edges based on Eq.~(\ref{eqn:eq7})
        \FOR{each layer in the convolution}
            \STATE Update $Z_s$ embedding based on Eq.~(\ref{eqn:eq9})
            \STATE Update $Z_v$ embedding based on Eq.~(\ref{eqn:eq10})
        \ENDFOR
        \STATE Read out the graph embedding based on Eq.~(\ref{eqn:eq11})
        \STATE Optimize the loss function based on Eq.~(\ref{eqn:eq12})
    \ENDFOR
\ENDFOR
\STATE \textbf{return} Predicted labels for the video clips and sequence data segments
\end{algorithmic}
\end{algorithm}

\section{Experiments}

\begin{table}[b]
\centering
\renewcommand{\arraystretch}{1.10}
\begin{tabular}{ccccc}
\toprule
{Datasets}   & {AudioSet} & {AVE}  & {RAVDESS} & {FMF}  \\
\midrule
Training   & 59840    & 3334 & 1118    & 4250 \\
Validation & 17097    & 402  & 160     & 0    \\
Test       & 8550     & 402  & 320     & 1053 \\
\midrule
Total      & 85487    & 4138 & 1598    & 5303 \\
\bottomrule
\end{tabular}
\caption{Statistics of the four multimodal datasets.}
\label{table_datasets}
\end{table}

\begin{table*}[t!]
\centering
\begin{tabular}{c|c|cc|cc|cc|cc}
\hline
\multirow{2}{*}{Model }     & \multirow{2}{*}{Params} & \multicolumn{2}{c|}{AudioSet} & \multicolumn{2}{c|}{AVE} & \multicolumn{2}{c|}{RAVDESS} & \multicolumn{2}{c}{FMF} \\ \cline{3-10} 
                              &                         & mAP           & AUC           & mAP         & AUC        & mAP           & AUC          & mAP        & AUC        \\ \hline
\multicolumn{1}{c|}{R(2+1)D}  & 33.4M                   & 36.2±0.5      & 81.4±0.8      & 57.8±2.3    & 89.4±2.6   & 61.2±1.4      & 90.1±1.2     & 88.4±1.1   & 92.6±2.2   \\
\multicolumn{1}{c|}{Wav2Vet2} & 94.5M                   & 42.4±2.1      & 88.6±1.7      & 68.4±1.6    & 92.3±1.3   & 67.4±1.2      & 91.8±1.6     & 69.3±1.8   & 89.1±1.0   \\
\multicolumn{1}{c|}{TMC}      & 0.1M                    &               53.6±1.4&               94.1±1.0&             74.5±1.4&            95.6±1.2&               75.5±1.8&              94.7±1.3&            90.4±1.7&            93.2±1.6\\
\multicolumn{1}{c|}{GBDT-HSM}      & 63.5M                    &               52.5±2.3&               94.6±2.1&             72.3±2.6&            94.6±1.7&               76.3±1.9&              94.9±1.5&            93.4±2.1&            94.3±1.6\\

\multicolumn{1}{c|}{HGCN}     & 42.4M                   & 44.2±1.3      & 88.3±1.1      & 72.1±1.5    & 93.0±2.1   & 72.4±0.6      & 93.4±1.5     & 91.5±0.8   & 92.1±1.0   \\
\multicolumn{1}{c|}{VAED}     & 2.1M                    & 50.7±1.6      & 91.2±1.5      & 75.7±0.7    & 96.1±1.4   & 64.9±1.1      & 92.2±0.9     & 89.3±1.9   & 93.2±1.4   \\
\multicolumn{1}{c|}{TMac}     & 4.4M                    & 56.2±0.8      & 94.6±1.3      & 77.2±1.3    & 95.2±1.6   & 72.9±1.2      & 93.6±1.7     & 93.1±0.6   & 94.6±0.9   \\
\multicolumn{1}{c|}{Ours}     & 2.1M                    & \textbf{57.6±1.2}      & \textbf{96.5±1.2}      & \textbf{79.2±1.4}    & \textbf{96.8±1.1}   & \textbf{82.2±1.6}      & \textbf{96.4±0.8}    & \textbf{96.4±0.6}   & \textbf{96.3±1.8}   \\ \hline
\end{tabular}
\caption {Classification results of four types of multimodal data using different methods.}
\label{table_R_3}
\end{table*}

\begin{table*}[t!]
\centering
\renewcommand{\arraystretch}{1.10}
\begin{tabular}{c|ll|ll|ll|ll}
\hline
\multirow{2}{*}{Modal } & \multicolumn{2}{c|}{AudioSet}                      & \multicolumn{2}{c|}{AVE}                           & \multicolumn{2}{c|}{RAVDESS}                       & \multicolumn{2}{c}{FMF}                           \\ \cline{2-9} 
                          & \multicolumn{1}{c}{mAP} & \multicolumn{1}{c|}{AUC} & \multicolumn{1}{c}{mAP} & \multicolumn{1}{c|}{AUC} & \multicolumn{1}{c}{mAP} & \multicolumn{1}{c|}{AUC} & \multicolumn{1}{c}{mAP} & \multicolumn{1}{c}{AUC} \\ \hline
SeqData                   & 48.1±0.8                & 91.2±1.2                 & 33.1±1.4                & 87.9±1.5                 & 64.8±1.3                & 91.5±1.8                 & 58.6±1.3                & 61.9±1.5                \\
Video                     & 50.3±1.5                & 92.4±1.9                 & 38.2±1.0                & 88.6±1.2                 & 69.2±1.4                & 87.8±1.2                 & 88.5±1.1                & 89.8±2.0                \\
Combined                  & \textbf{57.6±1.2} & \textbf{96.5±1.5} & \textbf{79.2±1.4} & \textbf{96.8±1.1} & \textbf{82.8±1.6} & \textbf{96.4±0.8} & \textbf{96.4±0.6} & \textbf{96.3±1.8} \\ \hline
\end{tabular}
\caption {Ablation studies on different modalities across four datasets.}
\label{table_R_2}
\end{table*}

\begin{table*}[t!]
\centering
\renewcommand{\arraystretch}{1.10}
\begin{tabular}{c|ll|ll|ll|ll}
\hline
\multicolumn{1}{l|}{\multirow{2}{*}{Setting}} & \multicolumn{2}{c|}{AudioSet}                      & \multicolumn{2}{c|}{AVE}                           & \multicolumn{2}{c|}{RAVDESS}                       & \multicolumn{2}{c}{FMF}                           \\ \cline{2-9} 
\multicolumn{1}{l|}{}                          & \multicolumn{1}{c}{mAP} & \multicolumn{1}{c|}{AUC} & \multicolumn{1}{c}{mAP} & \multicolumn{1}{c|}{AUC} & \multicolumn{1}{c}{mAP} & \multicolumn{1}{c|}{AUC} & \multicolumn{1}{c}{mAP} & \multicolumn{1}{c}{AUC} \\ \hline
w/o IM                                         & 21.2±1.1                & 78.1±2.3                 & 22.6±1.5                & 81.2±1.2                 & 51.2±2.1                & 81.5±1.8                 & 85.2±0.8                & 87.5±1.5                \\
w/o HG                                         & 56.2±0.8                & 94.6±1.3                 & 77.2±1.3                & 95.2±1.6                 & 72.9±1.2                & 93.6±1.7                 & 93.1±0.6                & 94.6±0.9                \\
Ours                                           & \textbf{57.6±1.2}                & \textbf{96.5±1.5}                 & \textbf{79.2±1.4}                & \textbf{96.8±1.1}                 & \textbf{82.8±1.6}                & \textbf{96.4±0.8}                 & \textbf{96.4±0.6}                & \textbf{96.3±1.8}                \\ \hline

\end{tabular}
\caption{Ablation studies on different methods across four datasets.}
\label{table_R_1}
\end{table*}

We conducted comprehensive experiments on four multimodal datasets: AudioSet~\cite{gemmeke2017audio}, AVE~\cite{tian2018ave}, RAVDESS~\cite{livingstone2018ryerson}, and FMF~\cite{wu2024cross}, covering baseline comparisons, training configurations, and parameter settings. To further evaluate model effectiveness, we performed ablation studies and hyperparameter analysis, examining the roles of temporal information, hypergraph structure, and the contributions of intra-modal and inter-modal components.

\subsection{Datasets}

\textbf{AudioSet:} includes 10-second YouTube video clips annotated with specific audio categories, widely used for audio-visual learning. We selected 33 high-confidence categories, resulting in 82,410 training clips. The evaluation set contains 85,487 test clips. The dataset is split into 70\% for training, 10\% for validation, and 20\% for testing.

\textbf{AVE:} is designed for multimodal learning tasks involving audio and visual events. It comprises 28 event categories, each sample containing a 10-second video and audio. The dataset is split into a training set of 3334 samples, a validation set of 402 samples, and a test set of 402 samples.

\textbf{RAVDESS:} recorded by 24 actors, covers 7 emotions in speech and song, and is commonly used for emotion recognition tasks. It contains 1598 audio and video samples, each 4 seconds long, with 70\% for training, 10\% for validation, and 20\% for testing.

\textbf{FMF: }The cross-modal dataset for anomaly detection in industrial processes contains 3 hours of synchronized video and three-phase AC data from various production batches. After splitting, the dataset is divided into 4250 10-second segments of training video, three-phase alternating current data, and 1053 test segments.

\subsection{Baselines}

\textbf{R(2+1)D}~\cite{tran2018closer}\textbf{:} A spatiotemporal method that splits 3D convolutions into 2D spatial and 1D temporal operations for video-based action recognition.

\textbf{Wav2Vec2}~\cite{baevski2020wav2vec}\textbf{:} A self-supervised model for audio recognition that learns from raw audio by predicting masked segments.

\textbf{GBDT-HSM}~\cite{xu2025novel}\textbf{:} A novel heterogeneous data classification approach combining gradient boosting decision trees and a hybrid structure model.

\textbf{TMC}~\cite{han2022trusted}\textbf{:} A novel multi-view classification algorithm that dynamically integrates different views at the evidence level.

\textbf{TMac}~\cite{liu2023tmac}\textbf{:} A well-designed graph convolutional network for temporal multi-modality aggregation.

\textbf{VAED}~\cite{shirian2022visually}\textbf{:} A multimodal framework that uses heterogeneous graphs to model the interactions between different modalities.

\textbf{HGCN}~\cite{shirian2023heterogeneous}\textbf{:} An advanced extension of VAED that enhances cross-modal learning through heterogeneous graph networks.

\subsection{Settings}

We conduct experiments on four datasets comprising 10-second and 4-second video segments with corresponding sequence data. A time-multimodal hybrid graph with 101 temporal nodes and 40 video nodes is constructed. Each experiment is repeated 10 times with different random seeds, and the mean average precision (mAP) and area under the ROC curve (AUC) are reported. The HHN model consists of two layers integrating HGNNs and GATs, where HGNNs capture intra-modal high-order relations and GATs fuse cross-modal information. In the intra-modal setting, the hyperedge length is set to 4 with a hop interval of 2. Training is performed for 10,000 iterations (including 1,000 warm-up) with a batch size of 128, using the Adam optimizer with learning rates of 0.005, 0.001, and 0.01, decaying by 0.1 after 250 iterations. All experiments are conducted on NVIDIA Tesla V100 GPUs.

\begin{figure}[t]
\centering

\includegraphics[width=1.00\columnwidth]{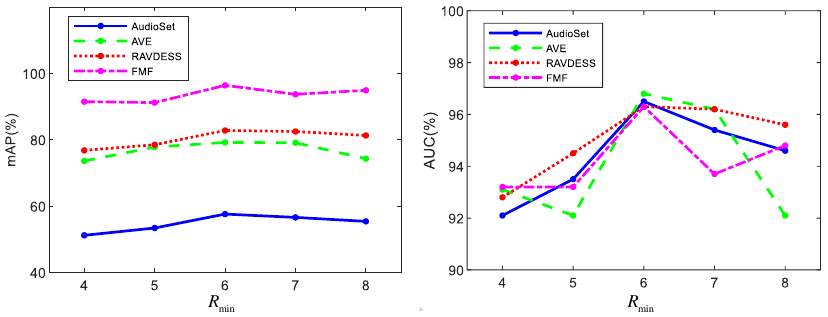}
\caption{
The effect of varying $R_{\min}$ values on mAP and AUC.}
\label{fig44}
\end{figure}

\subsection{Results}

As shown in Table~\ref{table_R_3}, the proposed HHN achieves SOTA performance across four datasets: AudioSet, AVE, RAVDESS, and FMF. Competing methods include video and audio models, a Bayesian uncertainty approach, and three graph-based models: TMac, VAED, and HGCN. HHN consistently outperforms all baselines. Specifically, AudioSet, AVE, and RAVDESS are audio-visual classification tasks, while FMF addresses industrial fault diagnosis using current and audio modalities. On AudioSet, HHN achieves an mAP of 57.6\% and an AUC of 96.5\%, surpassing TMac by 2.5\% and 1.9\%, respectively. On AVE and RAVDESS, HHN outperforms the second-best models in mAP by 2.0\% and 5.9\%, respectively. Its strong performance on FMF further demonstrates its generalization ability across diverse multimodal tasks. Notably, HHN attains SOTA results with only 2.1M parameters, highlighting its superior efficiency and compactness compared to larger graph-based models such as TMac, VAED, and HGCN.

\subsection{Hyperparameter Analysis}

We analyze the key hyperparameter $R_{\min}$ involved in the adaptive window function introduced in Section~\ref{sec:infra_modal_hypergraph}, which determines the minimum neighborhood size for entropy-guided hyperedge construction. As discussed, this parameter influences how local node entropy is translated into neighborhood scope, thereby affecting the expressiveness of intra-modal hypergraphs.

To evaluate its impact, we sweep $R_{\min} \!\in\! \{4,5,6,7,8\}$ across four datasets as shown in Figure \ref{fig44}. We observe that setting $R_{\min}{=}6$ yields the best overall performance. A smaller $R_{\min}$ tends to include weak or noisy connections, while a larger value may lead to excessive sparsification and loss of important local structures. The choice of $R_{\min}=6$ strikes a favorable balance between relational coverage and graph compactness.

\subsection{Ablation Study}

We conduct ablations from three perspectives.

\paragraph{Module ablation.}
As shown in Table \ref{table_R_1}, Ours outperforms both w/o HG (removing hypergraph convolution and using only traditional TGNNs for intra-/inter-modal propagation) and w/o IM (removing intra-modal convolution and keeping only inter-modal propagation) on AudioSet, AVE, RAVDESS, and FMF. This indicates that intra-modal hypergraphs and inter-modal temporal information are both necessary, and their combination is crucial to HHN's gains.

\paragraph{Modality ablation.}
Table \ref{table_R_2} studies modality contributions: SeqData uses only audio/current-sequence data, Video uses only the video modality, and Combined fuses audio and video. Ours (Combined) outperforms single-modality variants on every dataset, further validating the benefit of multimodal fusion.

\begin{figure}[t]
\centering
\hspace*{2pt} 
\includegraphics[width=1.00\columnwidth]{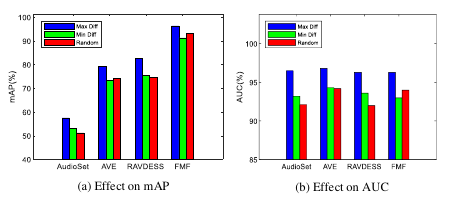}
\caption{
The effect of different entropy-based hyperedge construction methods on the mAP and AUC.
}
\label{fig45}
\end{figure}

\paragraph{Construction-strategy ablation.}
Under the setting $L=4$ and $R_{\min}=6$, we compare three ways to select hyperedge neighbors as shown in Figure \ref{fig45}:

\begin{itemize}[itemsep=2pt,topsep=2pt,parsep=0pt]
\item \textbf{Max Diff} -- choose nodes with the largest entropy differences;
\item \textbf{Min Diff} -- choose nodes with the smallest entropy differences;
\item \textbf{Random} -- choose nodes uniformly at random.
\end{itemize}

Experimental results show that ``Max Diff'' consistently achieves the best performance across all datasets, confirming its effectiveness for hybrid hypergraph construction.

\section{Conclusion}


This paper highlights the importance of temporal information in multimodal classification tasks. Sequential and video data with inherent temporal order are divided into segments as multimodal nodes, and a hybrid hypergraph is constructed to capture high-order intra-modal relations and inter-modal temporal dependencies. The proposed method integrates HGNNs and GATs, and optimizes hyperedge selection via adaptive window functions and maximum entropy difference. Experimental results demonstrate the superiority and reliability of our HHN model, confirming the role of intra-modal temporal information and high-order structural relations. In the future, we will further explore integrating temporal dynamics and multimodal learning to improve model adaptability, robustness, and generalization.

\bigskip

\bibliography{aaai2026}

\end{document}